# Counterfactual Reasoning in Linear Structural Equation Models

**Zhihong Cai**
Department of Biostatistics
Graduate School of Public Health
Kyoto University
Konoe-cho,Yoshida,Sakyo-ku,Kyoto,Japan
cai@pbh.med.kyoto-u.ac.jp

**Manabu Kuroki**
Department of Systems Innovation
Graduate School of Engineering Science
Osaka University
1-3,Machikaneyama-cho,Toyonaka,Osaka,Japan
mkuroki@sigmath.es.osaka-u.ac.jp

## Abstract

Consider the case where causal relations among variables can be described as a Gaussian linear structural equation model. This paper deals with the problem of clarifying how the variance of a response variable would have changed if a treatment variable were assigned to some value (counterfactually), given that a set of variables is observed (actually). In order to achieve this aim, we reformulate the formulas of the counterfactual distribution proposed by Balke and Pearl (1995) through both the total effects and a covariance matrix of observed variables. We further extend the framework of Balke and Pearl (1995) from point observations to interval observations, and from an unconditional plan to a conditional plan. The results of this paper enable us to clarify the properties of counterfactual distribution and establish an optimal plan.

## 1 INTRODUCTION

Causal inference with counterfactual reasoning is widely used in epidemiology, economics, and political science. An example of counterfactual is that "if I had taken aspirin, my headache would have gone now", which implies that in the actual world I did not take aspirin, and I still have the headache now. This example compares two outcomes: the actual outcome that I have the headache now because I did not take aspirin, and the counterfactual outcome that my headache would have gone if I had taken aspirin.

Evaluation of counterfactual queries plays an important role in treatment estimation, lawsuit compensation on hazardous exposure, planning and policy analysis. In medical science, comparison of the actual outcome and the counterfactual outcome is important to estimate the treatment or exposure effect in clinical trials and epidemiological studies. In economics, in order to evaluate the merit of a policy (e.g., taxation), all the possible influences in various counterfactual worlds are compared, where each world is created by a hypothetical implementation of a policy.

Counterfactual reasoning has been studied by many researchers in epidemiology (e.g. Greenland and Robins, 1988; Robins, 2004; Robins and Greenland, 1989a, 1989b). It is also one of the hot topics in artificial intelligence (e.g. Pearl, 1999, 2000; Tian and Pearl, 2000a, 2000b). Balke and Pearl (1994a, 1994b, 1995) presented formal notation, semantics and inference algorithms for the evaluation of counterfactual queries, and provided computational methods for the counterfactual distribution in the context of structural equation models.

In this paper, we examine counterfactual queries problem in the framework of linear structural equation models. We assume that causal knowledge is specified by linear structural equation models. Then, given that a set of variables is observed in the actual world, the aim of this paper is to examine how the mean and variance of a response variable would have changed if a treatment variable were assigned to some other value in the counterfactual world. Balke and Pearl (1994a, 1995) provided counterfactual formulas based on the given distribution of disturbances, but in general it may be difficult to know the distribution of disturbances. Therefore, in this paper, we consider using the variables in a path diagram to represent the distribution of disturbances. Then we formulate the mean and variance of the counterfactual distribution through both the total effect of a treatment variable on a response variable and the covariance matrix of a set of observed variables. Based on this formulation, we can decide the variables needed to be observed in order to examine the change of the mean and variance of a response variable in the counterfactual world. It is shown that for this purpose we need not observe all the variables in the path diagram, but only a subset on it. This formulation enables us to clarify the properties of counterfactual distribution.

This paper first considers the case where a set of point observations is available. Then we use the observed data to update the distribution of disturbances. We then evaluate the counterfactual mean and variance of a response variable if a fixed intervention of the treatment variable $X = x_0$ were conducted, which is called an unconditional plan. By reformulating the formulas proposed by Balke and Pearl (1995), we represent the counterfactual distribution through both the total effect and a covariance matrix of observed variables, which makes it easier to analyze the properties of counterfactual distribution. Second, we consider the case where a set of interval observations $r_1 \leq R \leq r_2$ are observed. We update distribution of disturbances by using the interval observations. Then we evaluate how the mean and variance of a response variable would have changed if a conditional plan $X = x_0 + aW$ were conducted, which means that the value of $X$ was determined by a set of observed variables $W$. Through this formula, we can select an optimal plan that minimizes the variance of the response variable.

## 2 LINEAR STRUCTRAL EQUATION MODEL

### 2.1 PATH DIAGRAM

In statistical causal analysis, a directed acyclic graph that represents cause-effect relationships is called a path diagram. A directed graph is a pair $G = (V, E)$, where $V$ is a finite set of vertices and the set $E$ of arrows is a subset of the set $V \times V$ of ordered pairs of distinct vertices. Regarding the graph theoretic terminology used in this paper, for example, refer to Kuroki and Cai (2004).

### DEFINITION 1 (PATH DIAGRAM)

Suppose a directed acyclic graph $G = (V, E)$ with a set $V = \{V_1, \cdots, V_n\}$ of variables is given. The graph $G$ is called a path diagram, when each child-parent family in the graph $G$ represents a linear structural equation model

$$V_i = \sum_{V_j \in \mathrm{pa}(V_i)} \alpha_{v_i v_j} V_j + \epsilon_{v_i}, \qquad i = 1, \ldots, n, \quad (1)$$

where $\mathrm{pa}(V_i)$ denotes a set of parents of $V_i$ in $G$ and $\epsilon_{v_1}, \ldots, \epsilon_{v_n}$ are assumed to be normally distributed. In addition, $\alpha_{v_i v_j} (\neq 0)$ is called a path coefficient. □

For detailed discussion regarding linear structural equation models, refer to Bollen (1989).

Here, we define some notations for future use. Let $\sigma_{xy \cdot z} = \mathrm{cov}(X, Y|Z)$, $\sigma_{yy \cdot z} = \mathrm{var}(Y|Z)$. In addition, $\beta_{yx \cdot z} = \sigma_{xy \cdot z}/\sigma_{xx \cdot z}$ is a regression coefficient of $X$ in the regression model of $Y$ on $\{x\} \cup z$. When $Z$ is an empty set, $Z$ is omitted from these arguments. The similar notations are used for other parameters.

For a set $Z$ of variables not including descendants of $V_j$, if $Z$ d-separates $V_i$ from $V_j$ in the graph obtained by deleting from a graph $G$ an arrow pointing from $V_i$ to $V_j$, then $\beta_{v_j v_i \cdot z} = \alpha_{v_j v_i}$ holds true. This criterion is called "the single door criterion" (e.g. Pearl, 2000). In addition, when $Z$ d-separates $V_i$ from $V_j$ in the graph $G$, $V_i$ is conditionally independent of $V_j$ given $Z$ in the corresponding distribution (e.g. Pearl, 1988, 2000; Spirtes et al., 1993).

### 2.2 IDENTIFIABILITY CRITERIA FOR TOTAL EFFECTS

Given a path diagram $G$, we wish to evaluate total effects from the correlation parameters between variables, where a total effect $\tau_{yx}$ of $X$ on $Y$ is defined as the total sum of the products of the path coefficients on the sequence of arrows along all directed paths from $X$ to $Y$. However, in many cases, it is difficult to obtain all the correlation parameters, since there usually exist unobserved variables. Hence, it is important to recognize sufficient sets of observed variables in order to evaluate total effects. Pearl (2000), Brito (2003), Brito and Pearl (2002a, 2002b, 2002c) and Tian (2004) provided identifiability criteria for causal parameters such as total effects, where "identifiable" means that causal parameters can be estimated consistently. In this paper, we introduce the back door criterion (e.g. Pearl, 2000) and the conditional instrumental variable method (Brito and Pearl, 2002a) as graphical identifiability criteria for total effects.

### DEFINITION 2 (BACK DOOR CRITERION)

Let $\{X, Y\}$ and $T$ be disjoint subsets of $V$ in a path diagram $G$. If a set $T$ of variables satisfies the following conditions relative to an ordered pair $(X, Y)$ of variables, then $T$ is said to satisfy the back door criterion relative to $(X, Y)$.

1. No vertex in $T$ is a descendant of $X$, and

2. $T$ d-separates $X$ from $Y$ in $G_{\underline{X}}$,

where $G_{\underline{X}}$ is the graph obtained by deleting from a graph $G$ all arrows emerging from vertices in $X$. □

If a set $T$ of observed variables satisfies the back door criterion relative to $(X, Y)$ in a path diagram $G$, then the total effect $\tau_{yx}$ of $X$ on $Y$ is identifiable through the observation of $\{X, Y\} \cup T$, and is given by the formula $\beta_{yx \cdot t}$ (Pearl, 2000).

**DEFINITION 3 (CONDITIONAL INSTRUMENTAL VARIABLE (IV))**

Let $\{X, Y, Z\}$ and $\boldsymbol{T}$ be disjoint subsets of $\boldsymbol{V}$ in a path diagram $G$. If a set $\boldsymbol{T} \cup \{Z\}$ of variables satisfies the following conditions relative to an ordered pair $(X, Y)$ of variables, then $Z$ is said to be a conditional instrumental variable (IV) given $\boldsymbol{T}$ relative to $(X, Y)$.

1. $\boldsymbol{T}$ is a subset of nondescendants of $Y$ in $G$,

2. $\boldsymbol{T}$ d-separates $Z$ from $Y$ but not from $X$ in $G_{\underline{X}}$. □

When $Z$ is a conditional instrumental variable given $\boldsymbol{T}$ relative to $(X, Y)$, a total effect of $X$ on $Y$ is identifiable through the observation of $\{X, Y, Z\} \cup \boldsymbol{T}$, and is given by $\sigma_{yz \cdot t}/\sigma_{xz \cdot t}$ (Brito and Pearl, 2002a).

Regarding the discussion about selection of identifiability criteria, refer to Kuroki and Cai (2004).

## 3 COUNTERFACTUAL ANALYSIS

### 3.1 REFORMULATION OF BALKE AND PEARL (1995)

In this section, we follow the counterfactual reasoning procedure proposed by Balke and Pearl (1995), and reformulate their formulas by representing counterfactual mean and variance through path coefficients and a covariance matrix of observed variables.

For this purpose, we partition a set $\boldsymbol{V}$ of vertices in a path diagram $G$ into the following three disjoint sets:

$\boldsymbol{S} = \{Y\} \cup \boldsymbol{U}$ : a set of descendants of $X$ whose first component is a response variable $Y$ of interest ($Y \notin \boldsymbol{U}$),

$X$ : a treatment variable,

$\boldsymbol{T} = \boldsymbol{Z} \cup \boldsymbol{W} = \boldsymbol{V} \setminus (\{X\} \cup \boldsymbol{S})$ : a set of nondescendants of $X$ ($\boldsymbol{W} \cap \boldsymbol{Z} = \phi$).

Denote $n_s$ as the number of elements in $\boldsymbol{S}$, and the similar notations are used for other numbers. According to the above partition of $\boldsymbol{V}$, let $A_{st}$ be a path coefficient matrix of $\boldsymbol{T}$ on $\boldsymbol{S}$ whose $(i, j)$ component is the path coefficient of $T_j$ on $S_i$ ($S_i \in \boldsymbol{S}, T_j \in \boldsymbol{T}$). Let $\boldsymbol{0}_{xs}$ be an $(n_x, n_s)$ zero matrix and $I_{ss}$ an $n_s$ dimensional identity matrix, respectively. The similar notations are used for other matrices.

Then, equation (1) can be rewritten as follows:

$$\begin{pmatrix} \boldsymbol{S} \\ X \\ \boldsymbol{T} \end{pmatrix} = \begin{pmatrix} A_{ss} & A_{sx} & A_{st} \\ \boldsymbol{0}_{xs} & 0 & A_{xt} \\ \boldsymbol{0}_{ts} & \boldsymbol{0}_{tx} & A_{tt} \end{pmatrix} \begin{pmatrix} \boldsymbol{S} \\ X \\ \boldsymbol{T} \end{pmatrix} + \begin{pmatrix} \boldsymbol{\epsilon}_s \\ \epsilon_x \\ \boldsymbol{\epsilon}_t \end{pmatrix}, \quad (2)$$

where $\boldsymbol{\epsilon}_s$, $\epsilon_x$ and $\boldsymbol{\epsilon}_t$ are random disturbance vectors corresponding to $\boldsymbol{S}$, $X$ and $\boldsymbol{T}$, respectively. In addition,

$$A_{sx} = \begin{pmatrix} A_{yx} \\ A_{ux} \end{pmatrix}, A_{tt} = \begin{pmatrix} A_{zz} & A_{zw} \\ A_{wz} & A_{ww} \end{pmatrix},$$

$$A_{st} = (A_{sz}, A_{sw}) = \begin{pmatrix} A_{yz} & A_{yw} \\ A_{uz} & A_{uw} \end{pmatrix},$$

and $A_{xt} = (A_{xz}, A_{xw})$.

Here, we define some notations for future use. For sets $\boldsymbol{X}$, $\boldsymbol{Y}$ and $\boldsymbol{Z}$, let $B_{yx \cdot z}$ be the regression coefficient matrix of $\boldsymbol{x}$ in the regression model of $\boldsymbol{Y}$ on $\boldsymbol{x} \cup \boldsymbol{z}$, and let $\Sigma_{xy \cdot z}$ be the conditional covariance matrix between $\boldsymbol{X}$ and $\boldsymbol{Y}$ given $\boldsymbol{Z}$. In addition, let $\Sigma_{xx \cdot z}$ be the conditional covariance matrix of $\boldsymbol{X}$ given $\boldsymbol{Z}$. Then, $B_{yx \cdot z}$ can be evaluated by $\Sigma_{yx \cdot z} \Sigma_{xx \cdot z}^{-1}$. Furthermore, let $\boldsymbol{\mu}_{y \cdot z}$ be the conditional mean vector of $\boldsymbol{Y}$ given $\boldsymbol{Z}$. Especially, when $\boldsymbol{Y}$ consists of one variable $Y$, the conditional mean of $Y$ given $\boldsymbol{Z}$ is denoted by $\mu_{y \cdot z}$. When $\boldsymbol{Z}$ is an empty set, $\boldsymbol{Z}$ is omitted from these arguments. The similar notations are used for other matrices and parameters.

#### 3.1.1 INTERVENTION

First, we evaluate the mean and variance of the response variable $Y$ when an external intervention $X = x_0$ is conducted. We use the variables in the path diagram and their path coefficients to represent the distribution of disturbances. From equation (2), the mean and the variance of $\boldsymbol{\epsilon}_v$ can be provided as

$$\begin{pmatrix} \boldsymbol{\mu}_{\epsilon_s} \\ \mu_{\epsilon_x} \\ \boldsymbol{\mu}_{\epsilon_t} \end{pmatrix} = \begin{pmatrix} I_{ss} - A_{ss} & -A_{sx} & -A_{st} \\ \boldsymbol{0}_{xs} & 1 & -A_{xt} \\ \boldsymbol{0}_{ts} & \boldsymbol{0}_{tx} & I_{tt} - A_{tt} \end{pmatrix} \begin{pmatrix} \boldsymbol{\mu}_s \\ \mu_x \\ \boldsymbol{\mu}_t \end{pmatrix}$$

and

$$\begin{pmatrix} \Sigma_{\epsilon_s \epsilon_s} & \Sigma_{\epsilon_s \epsilon_x} & \Sigma_{\epsilon_s \epsilon_t} \\ \Sigma_{\epsilon_x \epsilon_s} & \sigma_{\epsilon_x \epsilon_x} & \Sigma_{\epsilon_x \epsilon_t} \\ \Sigma_{\epsilon_t \epsilon_s} & \Sigma_{\epsilon_t \epsilon_x} & \Sigma_{\epsilon_t \epsilon_t} \end{pmatrix}$$

$$= \begin{pmatrix} I_{ss} - A_{ss} & -A_{sx} & -A_{st} \\ \boldsymbol{0}_{xs} & 1 & -A_{xt} \\ \boldsymbol{0}_{ts} & \boldsymbol{0}_{tx} & I_{tt} - A_{tt} \end{pmatrix}$$

$$\times \begin{pmatrix} \Sigma_{ss} & \Sigma_{sx} & \Sigma_{st} \\ \Sigma_{xs} & \sigma_{xx} & \Sigma_{xt} \\ \Sigma_{ts} & \Sigma_{tx} & \Sigma_{tt} \end{pmatrix}$$

$$\times \begin{pmatrix} I_{ss} - A_{ss} & -A_{sx} & -A_{st} \\ \boldsymbol{0}_{xs} & 1 & -A_{xt} \\ \boldsymbol{0}_{ts} & \boldsymbol{0}_{tx} & I_{tt} - A_{tt} \end{pmatrix}' \quad (3)$$

respectively.

Here, let $\boldsymbol{S}^*$ and $\boldsymbol{T}^*$ represent a set of descendants of $X$ and a set of nondescendants of $X$ after conducting an external intervention $X = x_0$ (The similar notations

are used for other discussions). Then, the modified structural equation model can be provided as

$$\begin{pmatrix} S^* \\ T^* \end{pmatrix} = \begin{pmatrix} A_{ss} & A_{st} \\ 0_{ts} & A_{tt} \end{pmatrix} \begin{pmatrix} S^* \\ T^* \end{pmatrix} + \begin{pmatrix} A_{sx} \\ 0_{tx} \end{pmatrix} x_0 + \begin{pmatrix} \epsilon_s \\ \epsilon_t \end{pmatrix}. \quad (4)$$

Let $\boldsymbol{\mu}_{s^*}$ and $\boldsymbol{\mu}_{t^*}$ be the mean vectors of $S^*$ and $T^*$, respectively. In addition, let $\Sigma_{s^*s^*}$, $\Sigma_{s^*t^*}$, and $\Sigma_{t^*t^*}$ be the covariance matrix of $S^*$, the covariance matrix between $S^*$ and $T^*$ and the covariance matrix of $T^*$, respectively. The similar notations are used for other parameters. Then, the mean vector and the covariance matrix of equation (4) are

$$\begin{pmatrix} \boldsymbol{\mu}_{s^*} \\ \boldsymbol{\mu}_{t^*} \end{pmatrix} = \begin{pmatrix} I_{ss} - A_{ss} & -A_{st} \\ 0_{ts} & I_{tt} - A_{tt} \end{pmatrix}^{-1}$$
$$\times \left( \begin{pmatrix} A_{sx} \\ 0_{tx} \end{pmatrix} x_0 + \begin{pmatrix} \boldsymbol{\mu}_{\epsilon_s} \\ \boldsymbol{\mu}_{\epsilon_t} \end{pmatrix} \right) \quad (5)$$

$$\begin{pmatrix} \Sigma_{s^*s^*} & \Sigma_{s^*t^*} \\ \Sigma_{t^*s^*} & \Sigma_{t^*t^*} \end{pmatrix} = \begin{pmatrix} I_{ss} - A_{ss} & -A_{st} \\ 0_{ts} & I_{tt} - A_{tt} \end{pmatrix}^{-1}$$
$$\times \begin{pmatrix} \Sigma_{\epsilon_s\epsilon_s} & \Sigma_{\epsilon_s\epsilon_t} \\ \Sigma_{\epsilon_t\epsilon_s} & \sigma_{\epsilon_t\epsilon_t} \end{pmatrix} \begin{pmatrix} I_{ss} - A_{ss} & -A_{st} \\ 0_{ts} & I_{tt} - A_{tt} \end{pmatrix}'^{-1} \quad (6)$$

respectively. Here, since

$$\begin{pmatrix} I_{ss} - A_{ss} & -A_{st} \\ 0_{ts} & I_{tt} - A_{tt} \end{pmatrix}^{-1}$$
$$\times \begin{pmatrix} I_{ss} - A_{ss} & -A_{st} & -A_{sx} \\ 0_{ts} & I_{tt} - A_{tt} & 0_{tx} \end{pmatrix}$$
$$= \begin{pmatrix} I_{ss} & 0_{st} & -(I_{ss} - A_{ss})^{-1} A_{sx} \\ 0_{ts} & I_{tt} & 0_{tx} \end{pmatrix}, \quad (7)$$

by substituting equation (3) for equation (6), we can obtain

$$\boldsymbol{\mu}_{s^*} = \boldsymbol{\mu}_s + \boldsymbol{\tau}_{sx}(x_0 - \mu_x) \quad (8)$$
$$\Sigma_{s^*s^*} = \Sigma_{ss} - \Sigma_{sx}\boldsymbol{\tau}'_{sx} - \boldsymbol{\tau}_{sx}\Sigma_{xs} + \boldsymbol{\tau}_{sx}\boldsymbol{\tau}'_{sx}\sigma_{xx}$$
$$= \Sigma_{ss \cdot x} + (\boldsymbol{\tau}_{sx} - B_{sx})(\boldsymbol{\tau}_{sx} - B_{sx})'\sigma_{xx}, (9)$$

where $\boldsymbol{\tau}_{sx} = (I_{ss} - A_{ss})^{-1} A_{sx}$.

Noting that the first component of $S$ is the response variable $Y$, then the mean and the variance of $Y$ when an external intervention is conducted are provided as

$$\mu_{y^*} = \mu_y + \tau_{yx}(x_0 - \mu_x), \quad (10)$$

and

$$\sigma_{y^*y^*} = \sigma_{yy \cdot x} + (\tau_{yx} - \beta_{yx})^2 \sigma_{xx} \quad (11)$$

respectively. Equations (10) and (11) are dependent on the total effect $\tau_{yx}$, the variances $\sigma_{xx}$ and $\sigma_{yy}$ of $X$ and $Y$, and the covariance $\sigma_{xy}$ between $X$ and $Y$.

Thus, the graphical criteria for identifying total effects stated in section 2.2 (the back door criterion and the conditional IV method) can be used to identify the mean and the variance of $Y$ when conducting an external intervention.

### 3.1.2 INTERVENTION CONDITIONING ON OBSERVATIONS

Next, we evaluate the mean and variance of the response variable $Y$ if an external intervention were conducted in the counterfactual world, given that the actual point observations $\boldsymbol{R} = \boldsymbol{r}$ are observed. From equation (2), the conditional mean vector and the conditional covariance matrix of $\boldsymbol{\epsilon}_v$ given $\boldsymbol{R} = \boldsymbol{r}$ are

$$\begin{pmatrix} \boldsymbol{\mu}_{\epsilon_s \cdot r} \\ \mu_{\epsilon_x \cdot r} \\ \boldsymbol{\mu}_{\epsilon_t \cdot r} \end{pmatrix} = \begin{pmatrix} I_{ss} - A_{ss} & -A_{sx} & -A_{st} \\ 0_{xs} & 1 & -A_{xt} \\ 0_{ts} & 0_{tx} & I_{tt} - A_{tt} \end{pmatrix} \begin{pmatrix} \boldsymbol{\mu}_{s \cdot r} \\ \mu_{x \cdot r} \\ \boldsymbol{\mu}_{t \cdot r} \end{pmatrix}$$

and

$$\begin{pmatrix} \Sigma_{\epsilon_s\epsilon_s \cdot r} & \Sigma_{\epsilon_s\epsilon_x \cdot r} & \Sigma_{\epsilon_s\epsilon_t \cdot r} \\ \Sigma_{\epsilon_x\epsilon_s \cdot r} & \sigma_{\epsilon_x\epsilon_x \cdot r} & \Sigma_{\epsilon_x\epsilon_t \cdot r} \\ \Sigma_{\epsilon_t\epsilon_s \cdot r} & \Sigma_{\epsilon_t\epsilon_x \cdot r} & \Sigma_{\epsilon_t\epsilon_t \cdot r} \end{pmatrix}$$
$$= \begin{pmatrix} I_{ss} - A_{ss} & -A_{sx} & -A_{st} \\ 0_{xs} & 1 & -A_{xt} \\ 0_{ts} & 0_{tx} & I_{tt} - A_{tt} \end{pmatrix}$$
$$\times \begin{pmatrix} \Sigma_{ss \cdot r} & \Sigma_{sx \cdot r} & \Sigma_{st \cdot r} \\ \Sigma_{xs \cdot r} & \sigma_{xx \cdot r} & \Sigma_{xt \cdot r} \\ \Sigma_{ts \cdot r} & \Sigma_{tx \cdot r} & \Sigma_{tt \cdot r} \end{pmatrix}$$
$$\times \begin{pmatrix} I_{ss} - A_{ss} & -A_{sx} & -A_{st} \\ 0_{xs} & 1 & -A_{xt} \\ 0_{ts} & 0_{tx} & I_{tt} - A_{tt} \end{pmatrix}'.$$

Letting $(\boldsymbol{\epsilon}_{s \cdot r}, \boldsymbol{\epsilon}_{z \cdot r}, \boldsymbol{\epsilon}_{w \cdot r})$ be the updated disturbances with mean and covariance matrix above, since the modified structural equation model if an external intervention $X = x_0$ were conducted in the counterfactual world is

$$\begin{pmatrix} S^* \\ T^* \end{pmatrix} = \begin{pmatrix} A_{ss} & A_{st} \\ 0_{ts} & A_{tt} \end{pmatrix} \begin{pmatrix} S^* \\ T^* \end{pmatrix} + \begin{pmatrix} A_{sx} \\ 0_{tx} \end{pmatrix} x + \begin{pmatrix} \epsilon_{s \cdot r} \\ \epsilon_{t \cdot r} \end{pmatrix},$$

the mean vector and the covariance matrix are

$$\begin{pmatrix} \boldsymbol{\mu}_{s^*} \\ \boldsymbol{\mu}_{t^*} \end{pmatrix} = \begin{pmatrix} I_{ss} - A_{ss} & -A_{st} \\ 0_{ts} & I_{tt} - A_{tt} \end{pmatrix}^{-1}$$
$$\times \left( \begin{pmatrix} A_{sx} \\ 0_{tx} \end{pmatrix} x_0 + \begin{pmatrix} \boldsymbol{\mu}_{\epsilon_s \cdot r} \\ \boldsymbol{\mu}_{\epsilon_t \cdot r} \end{pmatrix} \right)$$

and

$$\begin{pmatrix} \Sigma_{s^*s^*} & \Sigma_{s^*t^*} \\ \Sigma_{t^*s^*} & \Sigma_{t^*t^*} \end{pmatrix} = \begin{pmatrix} I_{ss} - A_{ss} & -A_{st} \\ 0_{ts} & I_{tt} - A_{tt} \end{pmatrix}^{-1}$$

$$\times \begin{pmatrix} \Sigma_{\epsilon_s\epsilon_s\cdot r} & \Sigma_{\epsilon_s\epsilon_t\cdot r} \\ \Sigma_{\epsilon_t\epsilon_s\cdot r} & \Sigma_{\epsilon_t\epsilon_t\cdot r} \end{pmatrix} \begin{pmatrix} I_{ss} - A_{ss} & -A_{st} \\ \mathbf{0}_{ts} & I_{tt} - A_{tt} \end{pmatrix}'^{-1}$$

respectively. From equation (7), we can obtain

$$\begin{aligned}
\boldsymbol{\mu}_{s^*} &= \boldsymbol{\mu}_{s\cdot r} + \boldsymbol{\tau}_{sx}(x_0 - \mu_{x\cdot r}) \\
\Sigma_{s^*s^*} &= \Sigma_{ss\cdot r} - \Sigma_{sx\cdot r}\boldsymbol{\tau}'_{sx} - \boldsymbol{\tau}_{sx}\Sigma_{xs\cdot r} + \boldsymbol{\tau}_{sx}\boldsymbol{\tau}'_{sx}\sigma_{xx\cdot r} \\
&= \Sigma_{ss\cdot xr} + (\boldsymbol{\tau}_{sx} - B_{sx\cdot r})(\boldsymbol{\tau}_{sx} - B_{sx\cdot r})'\sigma_{xx\cdot r} \\
&= \Sigma_{ss\cdot x} + (\boldsymbol{\tau}_{sx} - B_{sx})(\boldsymbol{\tau}_{sx} - B_{sx})'\sigma_{xx} \\
&\quad - (B_{sr} - \boldsymbol{\tau}_{sx}B_{xr})\Sigma_{rr}(B_{sr} - \boldsymbol{\tau}_{sx}B_{xr})'.
\end{aligned}$$

Thus, given $\boldsymbol{R} = \boldsymbol{r}$, the mean and variance of $Y$ if an external intervention $X = x_0$ were conducted are evaluated as

$$\mu_{y^*} = \mu_{y\cdot r} + \tau_{yx}(x_0 - \mu_{x\cdot r}), \quad (12)$$

$$\begin{aligned}
\sigma_{y^*y^*} &= \sigma_{yy\cdot x} + (\tau_{yx} - \beta_{yx})^2\sigma_{xx} \\
&\quad - (B_{yr} - \tau_{yx}B_{xr})\Sigma_{rr}(B_{yr} - \tau_{yx}B_{xr})'. \quad (13)
\end{aligned}$$

respectively. The last term in equation (13) is the correlation between $\boldsymbol{R}$ and $Y$ excluding the correlation between $\boldsymbol{R}$ and $Y$ via $X$.

### 3.2 PROPERTIES

Based on the reformulations in section 3.1, we can derive the following properties:

(I) It can be seen from equations (10), (11), (12) and (13) that the identifiability condition for counterfactual mean and variance is the same as that for the total effect $\tau_{yx}$ of $X$ on $Y$ when both $X$ and $Y$ are observed. That is, since these equations are only dependent on the total effect $\tau_{yx}$, the graphical criteria for identifying total effects stated in section 2.2 (the back door criterion and the conditional IV method) can also be used to identify the counterfactual mean and variance.

(II) Regarding equations (11) and (13), the first term is the conditional variance of $Y$ given $X$, and the second term is the square of the spurious correlation between $X$ and $Y$. The two terms are not dependent on the selection of $\boldsymbol{R}$. On the other hand, the last term in equation (13) is dependent on $\boldsymbol{R}$.

(III) When there is no confounder, since $(\tau_{yx} - \beta_{yx})^2\sigma_{xx} = 0$ holds true, both equations (11) and (13) are smaller than both the actual variance of $Y$ and the conditional variance of $Y$ given $X$.

(IV) When $\boldsymbol{R}$ satisfies the back door criterion relative to $(X, Y)$, we can obtain

$$\begin{aligned}
\sigma_{y^*y^*} &= \sigma_{yy\cdot x} + B_{yr\cdot x}B_{rx}\sigma_{xx}B'_{rx}B'_{yr\cdot x} \\
&\quad - B_{yr\cdot x}\Sigma_{rr}B'_{yr\cdot x} \\
&= \sigma_{yy\cdot x} - B_{yr\cdot x}\Sigma_{rr\cdot x}B'_{yr\cdot x} \\
&= \sigma_{yy\cdot xr} \leq \sigma_{yy\cdot x} \leq \sigma_{yy}
\end{aligned}$$

from Lemma 1 in Kuroki and Cai (2004). That is, when $\boldsymbol{R}$ satisfies the back door criterion relative to $(X, Y)$, the counterfactual variance of $Y$ is always smaller than both the actual variance of $Y$ and the conditional variance of $Y$ given $X$.

(V) In the case where the total effect of $X$ on $Y$ and the spurious correlation between $X$ and $Y$ have different signs, the counterfactual variance of $Y$ may be larger than that of $Y$ in the actual world. For example, letting $\boldsymbol{R}$ be an empty set, when $\tau_{yx} \neq 0$ but $\beta_{yx} = 0$ holds true in equation (11),

$$\sigma_{y^*y^*} = \sigma_{yy\cdot x} + (\tau_{yx} - \beta_{yx})^2\sigma_{xx} = \sigma_{yy} + \tau_{yx}^2\sigma_{xx} \geq \sigma_{yy}.$$

This situation occurs in the case where the correlation between $X$ and $Y$ is small although the spurious correlation is large, which is often called the parametric cancellation (refer to Cox and Wermuth, 1996).

### 3.3 EXTENSION OF BALKE AND PEARL (1995)

Given that a set of point observations $\boldsymbol{R} = \boldsymbol{r}$ is observed, Balke and Pearl (1995) evaluated the counterfactual mean and variance of a response variable $Y$ if a fixed intervention of a treatment variable $X = x_0$ were conducted, which is called as an unconditional plan in this paper. In this section, we extend their framework in two aspects: from an unconditional plan to a conditional plan, and from point observations to interval observations.

Suppose that a set of interval observations $\boldsymbol{r_1} \leq \boldsymbol{R} \leq \boldsymbol{r_2}$ are observed in the actual world. Here, $\boldsymbol{r_1} \leq \boldsymbol{R} \leq \boldsymbol{r_2}$ indicates that $r_{1i} \leq R_i \leq r_{2i}$ holds true for any $r_{1i} \in \boldsymbol{r_1}$, $R_i \in \boldsymbol{R}$ and $r_{2i} \in \boldsymbol{r_2}$. In addition, $\boldsymbol{R}$ can include a treatment variable $X$ and/or a response variable $Y$. Then we consider that a conditional plan were conducted in the counterfactual world, which means that the value of $X$ is set according to the following function, where $\boldsymbol{W}$ is a set of observed variables of nondescendants of $X$:

$$X = x_0 + \boldsymbol{aW}, \quad (14)$$

where $x_0$ and $\boldsymbol{a}$ are a constant value and a constant vector, respectively. When $\boldsymbol{a}$ is a non-zero vector, equation (14) is called a conditional plan, otherwise it is called an unconditional plan (e.g. Pearl, 2000).

In this section, in order to extend the results of Balke and Pearl (1995), we define the following notations. Let $\sigma_{xy\cdot[z]} = \text{cov}(X, Y|\boldsymbol{z_1} \leq \boldsymbol{Z} \leq \boldsymbol{z_2})$, $\sigma_{yy\cdot[z]} = \text{var}(Y|\boldsymbol{z_1} \leq \boldsymbol{Z} \leq \boldsymbol{z_2})$ and $\mu_{y\cdot[z]} = E(Y|\boldsymbol{z_1} \leq \boldsymbol{Z} \leq \boldsymbol{z_2})$. For sets $\boldsymbol{X}$, $\boldsymbol{Y}$ and $\boldsymbol{Z}$, Let $\boldsymbol{\mu}_{y\cdot[z]}$, $\Sigma_{xy\cdot[z]}$ and $\Sigma_{yy\cdot[z]}$ be the conditional mean vector of $\boldsymbol{Y}$ given $\boldsymbol{z_1} \leq \boldsymbol{Z} \leq \boldsymbol{z_2}$, the conditional covariance matrix between $\boldsymbol{X}$ and $\boldsymbol{Y}$ given $\boldsymbol{z_1} \leq \boldsymbol{Z} \leq \boldsymbol{z_2}$ and the conditional covariance matrix of $\boldsymbol{Y}$ given $z_1 \leq Z \leq z_2$, respectively. When $Z$ is an empty set, $Z$ is omitted from these arguments. The similar notations are used for other matrices and parameters.

First, we update the distribution of disturbances by using the set of interval observations $r_1 \leq R \leq r_2$. The mean vector and covariance matrix are

$$\begin{pmatrix} \mu_{\epsilon_s \cdot [r]} \\ \mu_{\epsilon_x \cdot [r]} \\ \mu_{\epsilon_t \cdot [r]} \end{pmatrix} = \begin{pmatrix} I_{ss} - A_{ss} & -A_{sx} & -A_{st} \\ 0_{xs} & 1 & -A_{xt} \\ 0_{ts} & 0_{tx} & I_{tt} - A_{tt} \end{pmatrix}$$
$$\times \begin{pmatrix} \mu_{s \cdot [r]} \\ \mu_{x \cdot [r]} \\ \mu_{t \cdot [r]} \end{pmatrix} \quad (15)$$

and

$$\begin{pmatrix} \Sigma_{\epsilon_s \epsilon_s \cdot [r]} & \Sigma_{\epsilon_s \epsilon_x \cdot [r]} & \Sigma_{\epsilon_s \epsilon_t \cdot [r]} \\ \Sigma_{\epsilon_x \epsilon_s \cdot [r]} & \sigma_{\epsilon_x \epsilon_x \cdot [r]} & \Sigma_{\epsilon_x \epsilon_t \cdot [r]} \\ \Sigma_{\epsilon_t \epsilon_s \cdot [r]} & \Sigma_{\epsilon_t \epsilon_x \cdot [r]} & \Sigma_{\epsilon_t \epsilon_t \cdot [r]} \end{pmatrix}$$
$$= \begin{pmatrix} I_{ss} - A_{ss} & -A_{sx} & -A_{st} \\ 0_{xs} & 1 & -A_{xt} \\ 0_{ts} & 0_{tx} & I_{tt} - A_{tt} \end{pmatrix}$$
$$\times \begin{pmatrix} \Sigma_{ss \cdot [r]} & \Sigma_{sx \cdot [r]} & \Sigma_{st \cdot [r]} \\ \Sigma_{xs \cdot [r]} & \sigma_{xx \cdot [r]} & \Sigma_{xt \cdot [r]} \\ \Sigma_{ts \cdot [r]} & \Sigma_{tx \cdot [r]} & \Sigma_{tt \cdot [r]} \end{pmatrix}$$
$$\times \begin{pmatrix} I_{ss} - A_{ss} & -A_{sx} & -A_{st} \\ 0_{xs} & 1 & -A_{xt} \\ 0_{ts} & 0_{tx} & I_{tt} - A_{tt} \end{pmatrix}', \quad (16)$$

respectively. Here,

$$\Sigma_{tt \cdot [r]} = \begin{pmatrix} \Sigma_{zz \cdot [r]} & \Sigma_{zw \cdot [r]} \\ \Sigma_{wz \cdot [r]} & \Sigma_{ww \cdot [r]} \end{pmatrix},$$
$$\Sigma_{ss \cdot [r]} = \begin{pmatrix} \sigma_{yy \cdot [r]} & \Sigma_{yu \cdot [r]} \\ \Sigma_{uy \cdot [r]} & \Sigma_{uu \cdot [r]} \end{pmatrix},$$
$$\Sigma_{st \cdot [r]} = \begin{pmatrix} \sigma_{yz \cdot [r]} & \Sigma_{yw \cdot [r]} \\ \Sigma_{uz \cdot [r]} & \Sigma_{uw \cdot [r]} \end{pmatrix}.$$

Thus, when a conditional plan $X = x_0 + aW$ were conducted in the counterfactual world, we can obtain

$$\begin{pmatrix} S^* \\ Z^* \\ W^* \end{pmatrix} = \begin{pmatrix} A_{sx} \\ 0_{zx} \\ 0_{wx} \end{pmatrix} x_0 + \begin{pmatrix} \epsilon_{s \cdot [r]} \\ \epsilon_{z \cdot [r]} \\ \epsilon_{w \cdot [r]} \end{pmatrix}$$
$$+ \begin{pmatrix} A_{ss} & A_{sz} & A_{sw} + A_{sx}a' \\ 0_{zs} & A_{zz} & A_{zw} \\ 0_{ws} & A_{wz} & A_{ww} \end{pmatrix} \begin{pmatrix} S^* \\ Z^* \\ W^* \end{pmatrix} (17)$$

where $(\epsilon_{s \cdot [r]}, \epsilon_{z \cdot [r]}, \epsilon_{w \cdot [r]})$ has the mean vector as equation (15) and the covariance matrix as equation (16). Thus, the mean vector and the covariance matrix if a control plan $X = x_0 + aW$ given $r_1 \leq R \leq r_2$ were conducted in the counterfactual world are

$$\begin{pmatrix} \mu_{s^*} \\ \mu_{z^*} \\ \mu_{w^*} \end{pmatrix} = \begin{pmatrix} I_{ss} - A_{ss} & -A_{sz} & -A_{sw} - A_{sx}a \\ 0_{zs} & I_{zz} - A_{zz} & -A_{zw} \\ 0_{ws} & -A_{wz} & I_{ww} - A_{ww} \end{pmatrix}^{-1}$$
$$\times \left( \begin{pmatrix} A_{sx} \\ 0_{zx} \\ 0_{wx} \end{pmatrix} x_0 + \begin{pmatrix} \mu_{\epsilon_s \cdot [r]} \\ \mu_{\epsilon_x \cdot [r]} \\ \mu_{\epsilon_t \cdot [r]} \end{pmatrix} \right), \quad (18)$$

and

$$\begin{pmatrix} \Sigma_{s^*s^*} & \Sigma_{s^*z^*} & \Sigma_{s^*w^*} \\ \Sigma_{z^*s^*} & \Sigma_{z^*z^*} & \Sigma_{z^*w^*} \\ \Sigma_{w^*s^*} & \Sigma_{w^*z^*} & \Sigma_{w^*w^*} \end{pmatrix}$$
$$= \begin{pmatrix} I_{ss} - A_{ss} & -A_{sz} & -A_{sw} - A_{sx}a \\ 0_{zs} & I_{zz} - A_{zz} & -A_{zw} \\ 0_{ws} & -A_{wz} & I_{ww} - A_{ww} \end{pmatrix}^{-1}$$
$$\times \begin{pmatrix} \Sigma_{\epsilon_s \epsilon_s \cdot [r]} & \Sigma_{\epsilon_s \epsilon_x \cdot [r]} & \Sigma_{\epsilon_s \epsilon_t \cdot [r]} \\ \Sigma_{\epsilon_x \epsilon_s \cdot [r]} & \sigma_{\epsilon_x \epsilon_x \cdot [r]} & \Sigma_{\epsilon_x \epsilon_t \cdot [r]} \\ \Sigma_{\epsilon_t \epsilon_s \cdot [r]} & \Sigma_{\epsilon_t \epsilon_x \cdot [r]} & \Sigma_{\epsilon_t \epsilon_t \cdot [r]} \end{pmatrix}$$
$$\times \begin{pmatrix} I_{ss} - A_{ss} & -A_{sz} & -A_{sw} - A_{sx}a \\ 0_{zs} & I_{zz} - A_{zz} & -A_{zw} \\ 0_{ws} & -A_{wz} & I_{ww} - A_{ww} \end{pmatrix}'^{-1} (19)$$

respectively. Here, since

$$\begin{pmatrix} I_{ss} - A_{ss} & -A_{sz} & -A_{sw} - A_{sx}a \\ 0_{zs} & I_{zz} - A_{zz} & -A_{zw} \\ 0_{ws} & -A_{wz} & I_{ww} - A_{ww} \end{pmatrix}^{-1}$$
$$\times \begin{pmatrix} I_{ss} - A_{ss} & -A_{sz} & -A_{sw} & -A_{sx} \\ 0_{zs} & I_{zz} - A_{zz} & -A_{zw} & 0_{zx} \\ 0_{ws} & -A_{wz} & I_{ww} - A_{ww} & 0_{wx} \end{pmatrix}$$
$$= \begin{pmatrix} I_{ss} & C_{sz} & C_{sw} & -(I_{ss} - A_{ss})^{-1}A_{sx} \\ 0_{zs} & I_{zz} & 0_{zw} & 0_{zx} \\ 0_{ws} & 0_{wz} & I_{ww} & 0_{wx} \end{pmatrix},$$

where

$$(C_{sz}, C_{sw}) = (I_{ss} - A_{ss})^{-1}(A_{sz}, A_{sw} + A_{sx}a)$$
$$- (I_{ss} - A_{ss})^{-1}(A_{sz}, A_{sw})$$
$$= (I_{ss} - A_{ss})^{-1}(0_{sz}, A_{sx}a),$$

by substituting equation (16) for equation (19), we can obtain

$$\Sigma_{s^*s^*} = \Sigma_{ss \cdot [r]} + \tau_{sx} a \Sigma_{ww \cdot [r]} a' \tau'_{sx} + \tau_{sx} \sigma_{xx \cdot [r]} \tau'_{sx}$$
$$+ \tau_{sx} a \Sigma_{ws \cdot [r]} + \Sigma_{sw \cdot [r]} a' \tau'_{sx}$$
$$- \tau_{sx} \Sigma_{xs \cdot [r]} - \Sigma_{sx \cdot [r]} \tau'_{sx}$$
$$- \tau_{sx} a \Sigma_{wx \cdot [r]} \tau'_{sx} - \tau_{sx} \Sigma_{xw \cdot [r]} a' \tau'_{sx}$$
$$= \left( \Sigma_{ss \cdot [r]} - \frac{\Sigma_{sx \cdot [r]} \Sigma_{xs \cdot [r]}}{\sigma_{xx \cdot [r]}} \right)$$
$$+ \left( \tau_{sx} - \frac{\Sigma_{sx \cdot [r]}}{\sigma_{xx \cdot [r]}} \right) \left( \tau_{sx} - \frac{\Sigma_{sx \cdot [r]}}{\sigma_{xx \cdot [r]}} \right)' \sigma_{xx \cdot [r]}$$
$$+ (\tau_{sx} a + \Sigma_{sw \cdot [r]} \Sigma^{-1}_{ww \cdot [r]} - \tau_{sx} \Sigma_{xw \cdot [r]} \Sigma^{-1}_{ww \cdot [r]})$$
$$\times \Sigma_{ww \cdot [r]}$$
$$\times (\tau_{sx} a + \Sigma_{sw \cdot [r]} \Sigma^{-1}_{ww \cdot [r]} - \tau_{sx} \Sigma_{xw \cdot [r]} \Sigma^{-1}_{ww \cdot [r]})'$$
$$- (\Sigma_{sw \cdot [r]} \Sigma^{-1}_{ww \cdot [r]} - \tau_{sx} \Sigma_{xw \cdot [r]} \Sigma^{-1}_{ww \cdot [r]})$$
$$\times \Sigma_{ww \cdot [r]}$$
$$\times (\Sigma_{sw \cdot [r]} \Sigma^{-1}_{ww \cdot [r]} - \tau_{sx} \Sigma_{xw \cdot [r]} \Sigma^{-1}_{ww \cdot [r]})'. \quad (20)$$

It is seen that only the third term of equation (20) is dependent on $\boldsymbol{a}$. When $(\boldsymbol{\tau}_{sx}\boldsymbol{a} + \Sigma_{sw\cdot[r]}\Sigma_{ww\cdot[r]}^{-1} - \boldsymbol{\tau}_{sx}\Sigma_{xw\cdot[r]}\Sigma_{ww\cdot[r]}^{-1}) = \boldsymbol{0}$ holds true, letting the $\boldsymbol{a}$ to $\check{\boldsymbol{a}}$, the variance of $Y$ if the conditional plan $X = x_0 + \check{\boldsymbol{a}}\boldsymbol{W}$ were conducted in the counterfactual world is the smallest in the conditional plan $X = x_0 + \boldsymbol{a}\boldsymbol{W}$. This conditional plan is called an optimal plan in this paper. Regarding the detailed discussion of an optimal plan and its application, refer to Kuroki (2005). In this case,

$$\boldsymbol{\mu}_{s^*} = \boldsymbol{\mu}_{s\cdot[r]} + \boldsymbol{\tau}_{ss}(x_0 - \mu_{x\cdot[r]}) + (\boldsymbol{\tau}_{sx}\Sigma_{xw\cdot[r]}\Sigma_{ww\cdot[r]}^{-1} - \Sigma_{sw\cdot[r]}\Sigma_{ww\cdot[r]}^{-1})\boldsymbol{\mu}_{w\cdot[r]},$$

and

$$\Sigma_{s^*s^*} = \left(\Sigma_{ss\cdot[r]} - \frac{\Sigma_{sx\cdot[r]}\Sigma_{xs\cdot[r]}}{\sigma_{xx\cdot[r]}}\right) + \left(\boldsymbol{\tau}_{sx} - \frac{\Sigma_{sx\cdot[r]}}{\sigma_{xx\cdot[r]}}\right)\left(\boldsymbol{\tau}_{sx} - \frac{\Sigma_{sx\cdot[r]}}{\sigma_{xx\cdot[r]}}\right)' \sigma_{xx\cdot[r]} - (\Sigma_{sw\cdot[r]}\Sigma_{ww\cdot[r]}^{-1} - \boldsymbol{\tau}_{sx}\Sigma_{xw\cdot[r]}\Sigma_{ww\cdot[r]}^{-1})\Sigma_{ww\cdot[r]} \times (\Sigma_{sw\cdot[r]}\Sigma_{ww\cdot[r]}^{-1} - \boldsymbol{\tau}_{sx}\Sigma_{xw\cdot[r]}\Sigma_{ww\cdot[r]}^{-1})'. \quad (21)$$

Thus, the mean and the variance of the response variable $Y$ if the optimal conditional plan $X = x_0 + \check{\boldsymbol{a}}\boldsymbol{W}$ given $\boldsymbol{r}_1 \leq \boldsymbol{R} \leq \boldsymbol{r}_2$ are evaluated as

$$\mu_{y^*} = \mu_{y\cdot[r]} + \tau_{yx}(x_0 - \mu_{x\cdot[r]}) + (\tau_{yx}\Sigma_{xw\cdot[r]}\Sigma_{ww\cdot[r]}^{-1} - \Sigma_{yw\cdot[r]}\Sigma_{ww\cdot[r]}^{-1})\boldsymbol{\mu}_{w\cdot[r]}$$

and

$$\sigma_{y^*y^*} = \left(\sigma_{yy\cdot[r]} - \frac{\sigma_{yx\cdot[r]}^2}{\sigma_{xx\cdot[r]}}\right) + \left(\tau_{yx} - \frac{\sigma_{yx\cdot[r]}}{\sigma_{xx\cdot[r]}}\right)^2 \sigma_{xx\cdot[r]} - (\sigma_{yw\cdot[r]}\Sigma_{ww\cdot[r]}^{-1} - \tau_{yx}\Sigma_{xw\cdot[r]}\Sigma_{ww\cdot[r]}^{-1})\Sigma_{ww\cdot[r]} \times (\sigma_{yw\cdot[r]}\Sigma_{ww\cdot[r]}^{-1} - \tau_{yx}\Sigma_{xw\cdot[r]}\Sigma_{ww\cdot[r]}^{-1})', \quad (22)$$

respectively.

### 3.4 PROPERTIES

Based on the formulation in section 3.3, the following properties are derived:

(I) Setting $\boldsymbol{a}$ to $\boldsymbol{0}$, in the case where $\boldsymbol{R}$ is an empty set, equation (20) is consistent with equation (9). In the case where $\boldsymbol{R} = \boldsymbol{r}_1 = \boldsymbol{r}_2$ holds true, equation (20) is consistent with equation (9). Thus, equation (20) is the extension of Balke and Pearl (1995).

(II) When setting the $\boldsymbol{a}$ to $\boldsymbol{0}$ in equation (20), we can obtain the counterfactual formulas in the case where an unconditional plan given $\boldsymbol{r}_1 \leq \boldsymbol{R} \leq \boldsymbol{r}_2$ were conducted.

(III) Since we can obtain

$$(I_{ss} - A_{ss})^{-1}\Sigma_{\epsilon_s w\cdot[r]}$$
$$= \Sigma_{sw\cdot[r]} - \boldsymbol{\tau}_{sx}\Sigma_{xw\cdot[r]} - (I_{ss} - A_{ss})^{-1}A_{sw}\Sigma_{ww\cdot[r]}$$
$$- (I_{ss} - A_{ss})^{-1}A_{sz}\Sigma_{zw\cdot[r]}$$

based on the updated distribution of disturbances, we can obtain

$$\Sigma_{s^*w^*} = (I_{ss} - A_{ss})^{-1}A_{sz}\Sigma_{zw\cdot[r]}$$
$$+ ((I_{ss} - A_{ss})^{-1}A_{sw} + \boldsymbol{\tau}_{sx}\check{\boldsymbol{a}})\Sigma_{ww\cdot[r]}$$
$$+ (I_{ss} - A_{ss})^{-1}\Sigma_{\epsilon_s w\cdot[r]} = \boldsymbol{0} \quad (23)$$

from equation (17) and $(\boldsymbol{\tau}_{sx}\check{\boldsymbol{a}} + \Sigma_{sw\cdot[r]}\Sigma_{ww\cdot[r]}^{-1} - \boldsymbol{\tau}_{sx}\Sigma_{xw\cdot[r]}\Sigma_{ww\cdot[r]}^{-1}) = \boldsymbol{0}$. That is, the optimal plan can be also interpreted as the conditional plan which cancels the correlation between $\boldsymbol{S}$ and $\boldsymbol{W}$ given $\boldsymbol{r}_1 \leq \boldsymbol{R} \leq \boldsymbol{r}_2$.

(IV) Letting $\sigma_{y^*y^*}^{(1)}$ and $\sigma_{y^*y^*}^{(2)}$ be the counterfactual variance if an optimal plan $X = x_0 + \check{\boldsymbol{a}}_{w_1}\boldsymbol{W}_1$ and another optimal plan $X = x_0 + \check{\boldsymbol{a}}_{w_2}\boldsymbol{W}_2$ were conducted respectively, if the third term in equation (21)

$$(\sigma_{yw_1\cdot[r]}\Sigma_{w_1w_1\cdot[r]}^{-1} - \tau_{yx}\Sigma_{xw_1\cdot[r]}\Sigma_{w_1w_1\cdot[r]}^{-1})\Sigma_{w_1w_1\cdot[r]}$$
$$\times (\sigma_{yw_1\cdot[r]}\Sigma_{w_1w_1\cdot[r]}^{-1} - \tau_{yx}\Sigma_{xw_1\cdot[r]}\Sigma_{w_1w_1\cdot[r]}^{-1})'$$
$$\geq (\sigma_{yw_2\cdot[r]}\Sigma_{w_2w_2\cdot[r]}^{-1} - \tau_{yx}\Sigma_{xw_1\cdot[r]}\Sigma_{w_2w_2\cdot[r]}^{-1})\Sigma_{w_2w_2\cdot[r]}$$
$$\times (\sigma_{yw_2\cdot[r]}\Sigma_{w_2w_2\cdot[r]}^{-1} - \tau_{yx}\Sigma_{xw_2\cdot[r]}\Sigma_{w_2w_2\cdot[r]}^{-1})',$$

then $\sigma_{y^*y^*}^{(1)} \leq \sigma_{y^*y^*}^{(2)}$ holds true. This property provides a covariate selection criteria for minimizing the counterfactual variance of $Y$.

## 4 DISCUSSION

Counterfactual reasoning is an important issue in many practical science, yet its theory is less developed. This paper considered counterfactual problems when causal relations among variables can be described as a Gaussian linear structural equation model. We first reformulated the formulas proposed by Balke and Pearl (1995), which enables us to clarify the properties of counterfactual distribution. In addition, we extended the framework of Balke and Pearl (1995) in two aspects: from point observations to interval observations, and from unconditional plan to conditional plan. The results of this paper will promote the application and development of counterfactual reasoning theory.

Finally, we would like to point out some further works about this theory. First, the discussion of this paper is based on linear structural equation models, then a natural extension is nonparametric structural equation models, which may be of interest in a number of applications. Second, this paper evaluated the counterfactual distribution when an external intervention is conducted on a treatment variable, then extension to more than one treatment variables is also a future work. Third, the results of this paper are applicable to acyclic graph models, then corresponding theory to cyclic graph models is needed to be developed.

**ACKNOWLEGDEMENT**

Thanks go to Kazushi Maruo and Hiroki Motogaito of Osaka University for their helpful discussion on this paper. The comments of the reviewers on preliminary versions of this paper are also acknowledged. This research was supported by the College Women's Association of Japan, the Sumitomo Foundation, the Murata Overseas Scholarship Foundation and the Ministry of Education, Culture, Sports, Science and Technology of Japan.

**REFERENCES**

Balke, A. and Pearl, J. (1994a). Probabilistic Evaluation of Counterfactual Queries, *Proceedings of the 12th National Conference on Artificial Intelligence*, 230-237.

Balke, A. and Pearl, J.(1994b). Counterfactual Probabilities: Computational Methods, Bounds and Identifications, *Proceeding of the 10th Conference on Uncertainty in Artificial Intelligence*, 11-18.

Balke, A. and Pearl, J.(1995). Counterfactuals and Policy Analysis in Structural Models, *Proceeding of the 11th Conference on Uncertainty in Artificial Intelligence*, 46-54.

Bollen, K. A. (1989). *Structural Equations with Latent Variables*, John Wiley & Sons.

Brito, C. (2003). A New Approach to the Identification Problem, *Advances in Artificial Intelligence: The 16th Brazilian Symposium on Artificial Intelligence*, 41-51.

Brito, C. and Pearl, J. (2002a). Generalized Instrumental Variables, *Proceeding of the 18th Conference on Uncertainty in Artificial Intelligence*, 85-93.

Brito, C. and Pearl, J. (2002b). A Graphical Criterion for the Identification of Causal Effects in Linear Models, *Proceedings of the 18th National Conference on Artificial Intelligence*, 533-538.

Brito, C. and Pearl, J. (2002c). A new identification condition for recursive models with correlated errors, *Structural Equation Modeling*, **9**, 459-474.

Cox, D. R. and Wermuth, N. (1996). *Multivariate Dependencies: Models, Analysis and Interpretation*, Chapman & Hall.

Greenland, S. and Robins, J. M. (1988). Conceptual Problems in the definition and Interpretation of Attributable fractions. *American Journal of Epidemiology*, **128**, 1185-1197.

Kuroki, M. (2005). Selection of a control plan by using causal network in statistical process analysis. Submitted.

Kuroki, M. and Cai, Z. (2004). Selection of Identifiability Criteria for Total Effects by Using Path Diagrams. *Proceeding of the 20th Conference on Uncertainty in Artificial Intelligence*, 333-340.

Pearl, J. (1988). *Probabilistic reasoning in intelligence systems*, Morgan Kaufmann.

Pearl, J. (1999). Probabilities of Causation: Three Counterfactual Interpretations and Their Identification, *Synthese*, **121**, 93-149.

Pearl, J. (2000). *Causality: Models, Reasoning, and Inference*, Cambridge University Press.

Robins, J. M. (2004). Should Compensation Schemes be based on the Probability of Causation or Expected Years of Life Lost? *Journal of Law and Policy*, **12**, 537-548.

Robins, J. M. and Greenland, S. (1989a). Estimability and Estimation of Excess and Etiologic Fractions, *Statistics in Medicine*, **8**, 845-859.

Robins, J. M. and Greenland, S. (1989b). The Probability of Causation under a Stochastic Model for Individual Risk, *Biometrics*, **45**, 1125-1138.

Spirtes, P., Glymour, C., and Schienes, R. (1993). *Causation, Prediction, and Search*, Springer-Verlag.

Tian, J. (2004). Identifying Linear Causal Effects, *Proceedings the 19th National Conference on Artificial Intelligence*, 104-110.

Tian, J. and Pearl, J. (2000a). Probabilities of Causation: Bounds and Identification. *Annals of Mathematics and Artificial Intelligence*, **28**, 287-313.

Tian, J. and Pearl, J. (2000b). Probabilities of Causation: Bounds and Identification. *Proceedings of the 16th Conference on Uncertainty in Artificial Intelligence*, 589-598.